\documentclass{article}

\usepackage[english]{babel}

\usepackage[english]{babel}

\usepackage[letterpaper,top=2cm,bottom=2cm,left=3cm,right=3cm,marginparwidth=1.75cm]{geometry}

\usepackage{amsmath}
\usepackage{graphicx}
\usepackage[colorlinks=true, allcolors=blue]{hyperref}
\usepackage{natbib}
\usepackage{float}

\title{Meta Learning for Few-Shot
Medical Text Classification}
\usepackage{authblk}
\author[1]{Pankaj Sharma}
\author[1]{Minh Tran}
\author[1]{Imran Qureshi}
\affil{Stanford Center for Professional Development}
\begin{document}
\maketitle

\section{Extended Abstract}
Medical professionals frequently work in a data constrained setting to provide insights across a diverse demographic. A few medical observations, for instance, informs the diagnosis and treatment of a patient. This suggests a unique setting for meta-learning to learn models that can quickly adapt to new medical tasks and provide insights unattainable by other methods. We investigate the use of meta-learning and robustness techniques on a broad corpus of benchmark text and medical data. To do this, we developed new data pipelines, combined language models with meta-learning approaches, and extended existing meta-learning algorithms to minimize worst case loss. We find that meta-learning on text is a suitable framework for text-based data, providing better data efficiency and comparable performance to few-shot language models and can be successfully applied to medical note data. Furthermore, meta-learning models coupled with DRO can improve worst case loss across disease codes.\newline
 
 The first challenge was to validate the effectiveness of meta-learning on natural language. We did this by validating several approaches on the CLINC150 dataset, which is a corpus of intent snippets (e.g. "where is the phone" labelled as find\_phone). After trying approaches between both text encoders (e.g. RNN) and meta-learning algorithms (e.g. MANN), we validated that a BERT model to generate text embeddings combined with MAML/Prototypical Network provides near 100\% accuracy for the entire dataset (on 150-class, 3-shots). \newline
 
 Applying this approach to medical data, we used MIMIC-III, a critical care database with over 2 million notes. However, most notes were far longer than what any text encoder could process, and had many errors and redundancies so we created an end to end data pipeline that can extract data from the MIMIC III corpus to be used in meta-learning. Pre-processing of data required text processing steps (lower case, remove stop words, remove special characters, lemmatize) and using a summarizer to satisfy the 512 token requirement by BERT. Furthermore, we developed a dataloading strategy to construct tasks automatically from the embedded dataset.\newline
  
 We experimented with semi-rare, popular and random disease codes that were fed into a baseline ProtoNet with many positive results. For instance, across random disease codes, the models achieved 73\% accuracy for 10-way, 5-shot classification tasks which represents better accuracy and memory efficiency over a fine-tuned language model. Comparing the performance of ProtoNets on semi-rare, random and popular disease codes we found that the random codes performed better than the popular and semi-rare disease codes. This can be attributed to the skewed distribution of the number of notes per disease code.\newline
 
 Finally, we investigated approaches with distributionally robust optimization, a strategy to minimize worst case loss across specific groups. We modified the loss functions for both ProtoNet and MAML for our experiments. Our model variations included implementing DRO with or without group adjustments and $l_2$ regularization. We found that DRO combined with MAML does improve prediction and does account for distribution shift, with the conclusion that Meta-learning models coupled with distributionally robust optimization (with some variations) can yield fairer models across disease codes.

 \newpage

\begin{abstract}
 Medical professionals frequently work in a data constrained setting to provide insights across a unique demographic. A few medical observations, for instance, informs the diagnosis and treatment of a patient. This suggests a unique setting for meta-learning, a method to learn models quickly on new tasks, to provide insights unattainable by other methods. We investigate the use of meta-learning and robustness techniques on a broad corpus of benchmark text and medical data. To do this, we developed new data pipelines, combined language models with meta-learning approaches, and extended existing meta-learning algorithms to minimize worst case loss. We find that meta-learning on text is a suitable framework for text-based data, providing better data efficiency and comparable performance to few-shot language models and can be successfully applied to medical note data. Furthermore, meta-learning models coupled with DRO can improve worst case loss across disease codes.
\end{abstract}

\section{Introduction}

Medical professionals rely on limited information to derive insights for varied patient demographics. Often medical professionals rely on notes to document history of care, diagnose medical issues, and hand off cases to other doctors. As a result, medical notes contain valuable information such as clinical observations, medical history, treatment plans, and demographic information. Furthermore, notes are used in a variety of medical billing use cases for hospital revenue management.

Recent advances in natural language processing (NLP) have made strides in creating language models with near human level performance on language tasks \cite{fewshotlearners}. These advances have also been applied to medical tasks based on medical notes \cite{icdperformancebert}. However, current language model techniques require a large sample to fine tune even for a single task (e.g. 1000) and have a large memory and computational footprint. This is especially striking for emerging illnesses (e.g. COVID-19 and variants), where medical note data is scarce and the need for accurate classifications is pressing.

Meta-learning provides an excellent framework for tackling these low-resource classification problems. In this framework, a model representation is learned across multiple tasks. Tasks are defined via a Support set of examples (K-shot) where the representation can quickly adapt to the examples via updates to its hidden state. Afterwards, the model can predict on a "query" set of examples. As a result, meta-learning on medical note tasks provides a significant opportunity to create robust, yet flexible models that can be deployed in data constrained healthcare settings.

In this study, we focus on few shot classification of medical texts to investigate whether meta-learning strategies can accurately learn across rare disease classification tasks. The learned representations should adapt well to new disease types and provide accurate classification with a limited number of examples. We use a large pre-trained language model (e.g. ClinicalBERT) to generate representations to train a MAML model and Prototypical Network. Our note corpus comes from MIMIC-III ('Medical Information Mart for Intensive Care') which is based on patients records from Beth Israel Deaconess Medical Center in Boston, Massachusetts and are classified according to 9th version of the International Classification of Diseases. See Dataset Discussion for more.

There is a also the potential for bias in these models \cite{biasinmedicine}. Medical notes can be stratified by both the scribe's and the patient demographics (e.g. ethnicity, disease type, specialty, etc.). Given the critical decisions that rely on the outputs of these models and the potential harm that can result from biases, we also investigate meta-learning with Distributionally Robust Optimization (DRO) to understand performance bias towards any specific disease classification. See Methods for more.

\section{Methods}
\subsection{BERT Meta-Learning Model Architecture}
Our primary approach was to first tackle the general NLP meta-learning problem by meta-learning on textual representations on the CLINC150 dataset, an intent classification corpus, and then use the successful modelling strategies across CLINC150 towards the MIMIC III medical notes dataset. Our validated architecture consists of a large pre-trained language model, which generates textual representations, or embeddings, from raw text. With these representations, we automatically construct tasks into a MAML model and Prototypical Network to learn across these tasks. 

Our embeddings are retrieved using BERT (Bidirectional Encoder Representations from Transformers) and its variants \cite{bertoriginalpaper}, which are pre-trained language models which jointly condition on both left and right context in all layers. These models had achieved state of the art on NLP benchmarks in 2019, and more importantly, are effective in representing text as word vectors (in 768 dimensions) that can be used for downstream tasks. As an example, we can input a CLINC150 meaning\_of\_life entry "what do you think life is really about" into BERT and retrieve a vector $e_1 \in R^{768}$. This vector would be closer to $e_2 \in R^{768}$ generated from another meaning\_of\_life entry "why are humans on earth", but far from $e_3 \in R^{768}$  the find\_phone entry "help me find my phone please".

\begin{figure*}
  \includegraphics[width=1.0\textwidth]{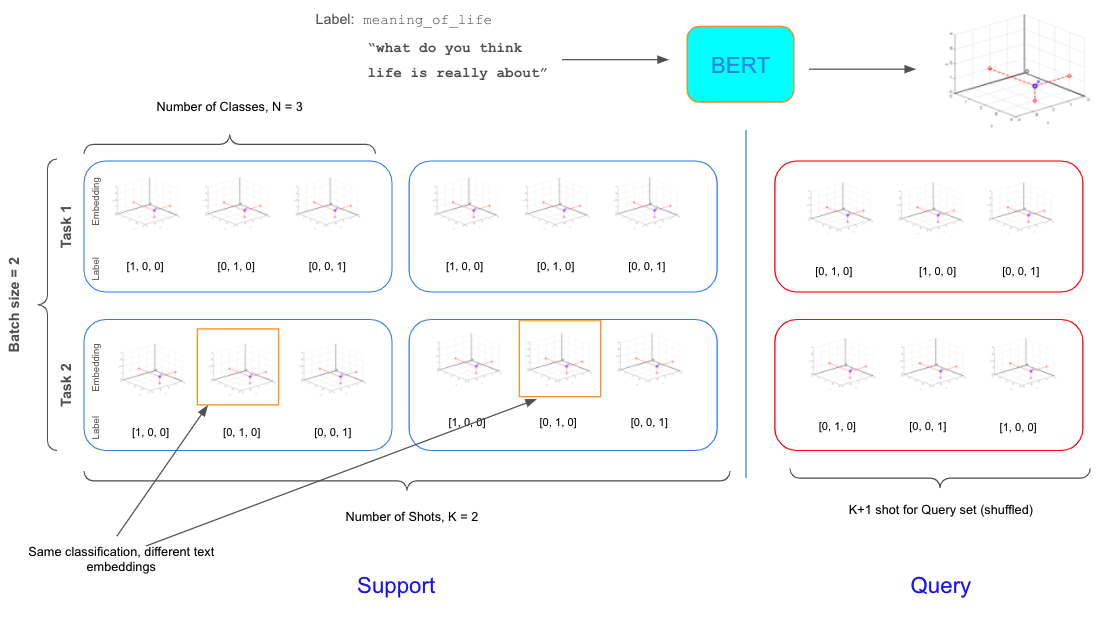}
  \caption{Data Loading Strategy: Create support and query sets from the word embeddings with N classes, K shots, and Q query size}
  \label {fig: Data Loading}
\end{figure*}

These embeddings are then constructed into tasks. Each task consists of a support set $D^{tr}$ with $K$ example across $N$ classes and a query set $D^{ts}$ with $Q$ query examples. See figure \ref{fig: Data Loading}. 

The tasks are used to train a meta-learning models: Model Agnostic Meta Learning model and Prototypical Network. Model Agnostic Meta Learning (MAML) \cite{finn2017modelagnostic} reformulates the training procedure into two steps, (1) an inner loop which is traditional optimization procedure across the task and (2) an outer loop to train across tasks. Combined these methods create a trained network that can easily adapt to multiple tasks.

Prototypical Networks (ProtoNets) also learn a representation across tasks,\cite{snell2017prototypical} but do so by learning a metric space where an encoder function $f_{\theta}$ projects raw examples into an embedding space. Instead of a set of weights learned from MAML, ProtoNets calculate prototypes $c_n$ which can be used to classify query examples via the $l_2$ distance from the query example to each class prototype:
\[
y^{ts} = \operatorname*{argmin}_n  \text{softmax} \space ||f_{\theta}(x^{ts}) - c_n||_2
\]
This approach provides a more memory efficient representation with an easier inductive bias as it is limited to classification problems, making ProtoNets an ideal model for our problem of classifying ICD codes from medical notes.

\begin{figure*}
  \includegraphics[width=1.0\textwidth]{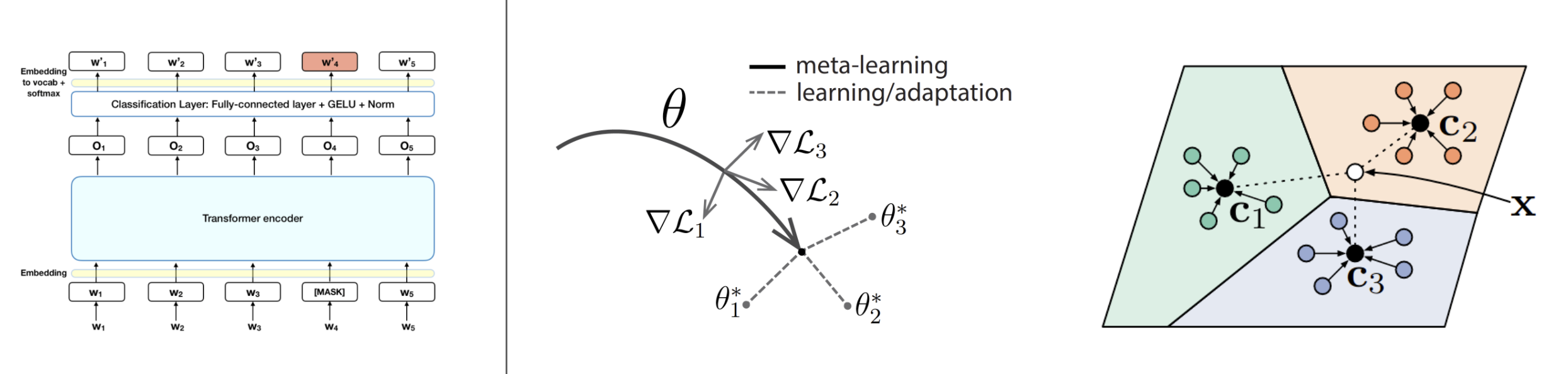}
  \caption{Experiment Architectures: Using embeddings for clinical notes using BERT (bidirectional encoder representations from transformers) and training using either MAML (including MAML with Distributionally Robust Optimization (DRO) in the adaptation loop) or ProtoNet. }
  \label {fig: Architecture}
\end{figure*}

\subsection{Distributionally Robust Optimization}

Finally, we investigate a distributionally robust loss function to our architecture. In Distributionally Robust Optimization (DRO) models are trained to minimize the worst case loss over a set of pre-defined groups. Shiori Sagaw et al. \cite{dronnforgroupshifts} showed that by strongly regularizing the the group DRO with $l_2$ regularization can give substantially better higher worst group accuracies. In our experiments we combined DRO and regularization with Meta-Learning to study impact on worst case group predictions (in this case, disease codes) during testing. The sampling was modified to include group codes (disease codes) in $D^{ts}$. These codes were used in the outer loop adaption for MAML to optimize for the worst case loss and track the worst performing groups.

\section{Datasets and Pre-Processing}
\subsection{Datasets} 
\begin{figure}[H]
  \centering
  \includegraphics[width=0.8\textwidth]{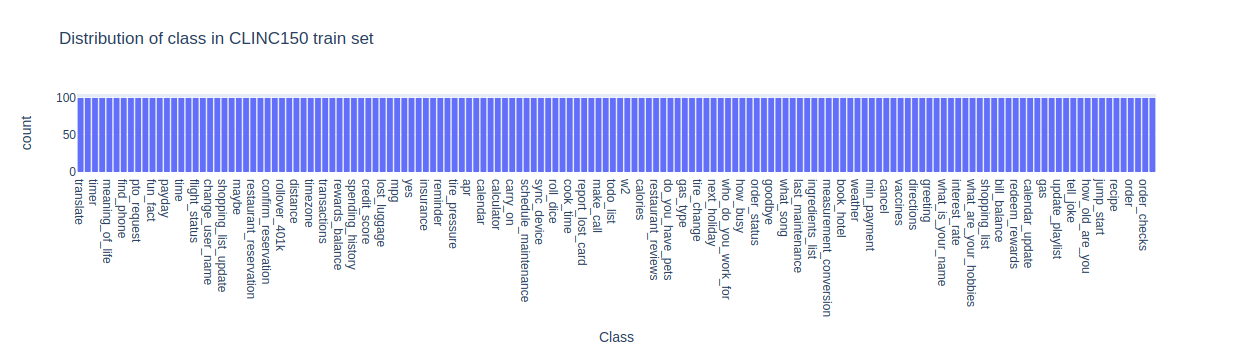}
  \caption{Intent Class Distribution of CLINC150 training set.}
  \label {fig: CLINC150 Class Dist.}
\end{figure}

Two datasets were used for experiements. The first CLINC150 dataset was used to establish a baseline to validate the meta-learning model. CLINC150 is used for Intent Classification and Out-of-Scope Prediction and is relatively simple \cite{fewshotclinc150}. The CLINC150 dataset was already processed and split into 6 sets including in/out of domain data for train, test, validation accordingly. For our baseline model, we used in domain data only, the data includes 150 "in-scope" intent classes, each with 100 records, 20 validations, and 30 test samples. Each record include a piece of text and its label (Figure \ref {fig: CLINC150 Sample}). The table below (Figure \ref {fig: CLINC150 Class Dist.}) shows the distribution of train and test set which is uniform distribution. The diversity of classes indicates that CLINC150 is a good dataset to establish a baseline, however the distribution of intent classes is much more constrained than the distribution of disease codes in MIMIC-III, where there are thousands of diseases and the most challenging to classify are the ones that are rare and often obscured by more common diseases.

\begin{figure}[H]
  \centering
  \includegraphics[width=0.8\textwidth]{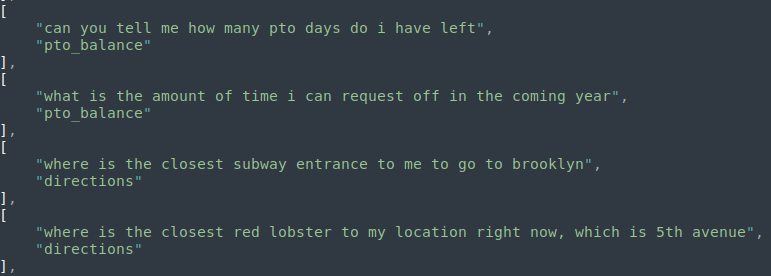}
  \caption{CLINC150 Sample Data}
  \label {fig:  CLINC150 Sample}
\end{figure}

\begin{figure}[H]
  \centering
  \includegraphics[width=0.8\textwidth]{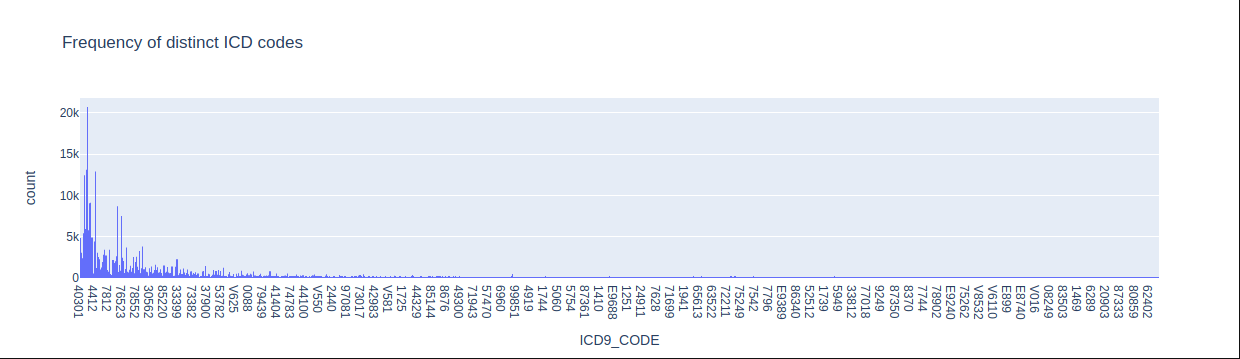}
  \caption{Original Data Distribution of all ICD Codes.}
  \label {fig: All ICD Codes}
\end{figure}

After establishing the baseline, we performed experiments using the MIMIC-III dataset, which is a large, freely-available database comprising de-identified health-related data associated with over 40,000 patients who stayed in critical care units of the Beth Israel Deaconess Medical Center between 2001 and 2012. For our scope, we extracted only specific data required for disease classification, which included patient IDs, associated medical notes across time, and associated diagnosis disease codes (or ICD codes). After selection, the dataset ended up containing around 2 millions records of more than 46,000 patients with 6,984 distinct diseases. See \ref{fig: All ICD Codes}. The distribution of codes is largely skewed towards heart- or respiratory-related diseases, with a long tail of rare disease codes.  Given time and processing constraints, we decided to take sampled 1000 records from each of the top 10 ICDs. We chose to equalize the ICD code samples so that the variance in distribution is due to properties of the medical notes themselves rather than the number of data samples. The selected data distribution is shown in (Figure \ref {fig: Top ICD}).

\subsection{Data Pre-Processing}
No data pre-processing was performed on the CLINC150 dataset as the data is clean and well organized. Unlike CLINC150, MIMIC-III medical notes contain many typos, errors, and abbreviations. Furthermore, many notes could be associated with a single patient across time, creating redundant information with too many tokens for our text encoder (BERT) to process into embeddings. 

\begin{figure}[H]
  \centering
  \includegraphics[width=0.8\textwidth]{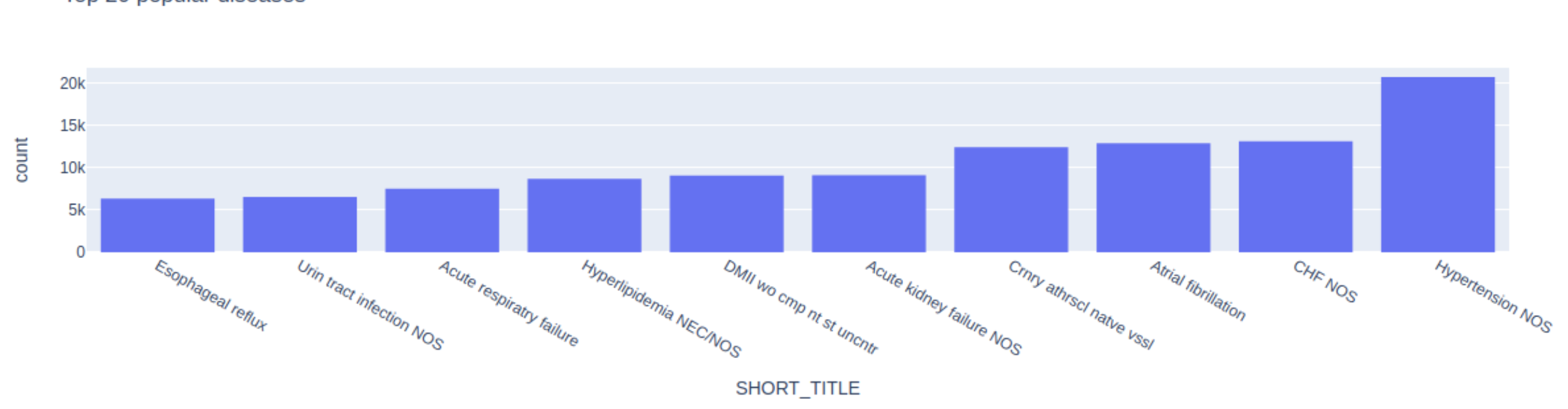}
  \caption{Top 10 most popular diseases.}
  \label {fig: Top ICD}
\end{figure}

As a result, we designed an end-to-end pipeline to extract and clean this MIMIC data. There were other efforts to process MIMIC-III text data \cite{2020mimicextract} \cite{nuthakki2019natural} for classification, we used these pipelines as references and built our own procedure. For data selection, as mentioned above, our pipeline extracts most popular ICD codes and most rare ICD codes from schema tables depending on the experiment settings, example of the extracted data distribution of top 10 ICD is shown in (Figure \ref {fig: Top ICD}). For each disease, we sampled 1000 medical notes while limiting the number of records from a same patient ID. By doing this, we can increase the variance of the data and minimize the redundant information since notes for a patient in a specific period usually contain same content with small updates.

\begin{figure}[H]
  \centering
  \includegraphics[width=0.8\textwidth]{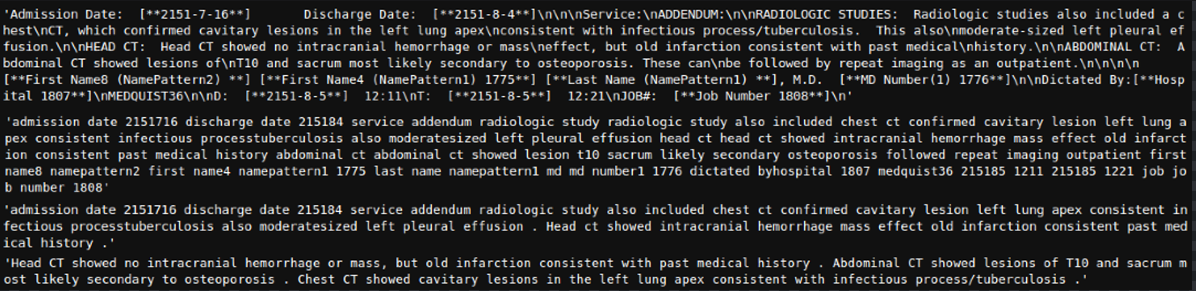}
  \caption{Step-by-step data processing.}
  \label {fig: Clean text}
\end{figure}

For each medical note, we applied the standard text processing steps (lower case, remove stop words, remove special characters, lemmatize) and finally solved the input length problem by summarizing the text into a string of maximum 512 tokens using BERT and BART Summarizer \cite{bertoriginalpaper} \cite{bartoriginalpaper}. The example below (Figure \ref {fig: Clean text}) demonstrates the processing steps for one specific example, the final text output will be used at input data for BERT to get the embedding vector. 

\section{Experiments, Results and Discussion}

\subsection{Experiment Architecture}
For experimentation (Figure \ref{fig: Architecture}) we generated embeddings from clinical notes using BERT (bidirectional encoder representations from transformers) and trained two meta-learning models: ProtoNets and MAML. We also extended MAML model by implementing a Distributionally Robust Optimization (DRO) to minimize the worst-case loss during adaptation.

\begin{figure}
  \centering
  \includegraphics[width=0.5\textwidth]{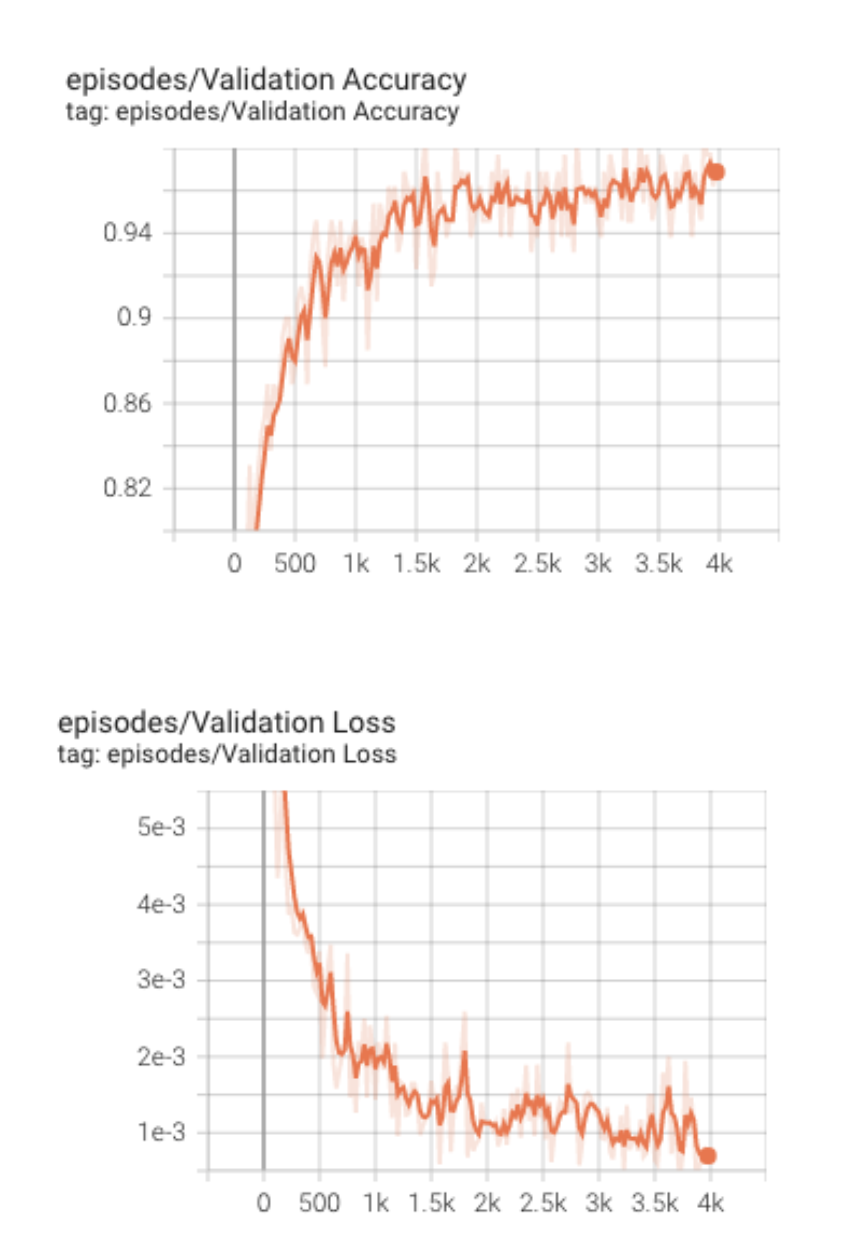}
  \caption{Baseline with ProtoNet using CLINC150 for a 150 class 3 shot model.}
  \label {fig: Protonet Baseline}
\end{figure}

The results of the baseline with CLINC150 using 512 token BERT embedding using the ProtoNet model is shown in Figure \ref{fig: Protonet Baseline}. For a 10 class 5 shot model, we achieved a meta accuracy of 98\% and for a 150 class 3 shot model, we achieved an accuracy of 96.7\% 

\begin{figure}
  \includegraphics[width=1.0\textwidth]{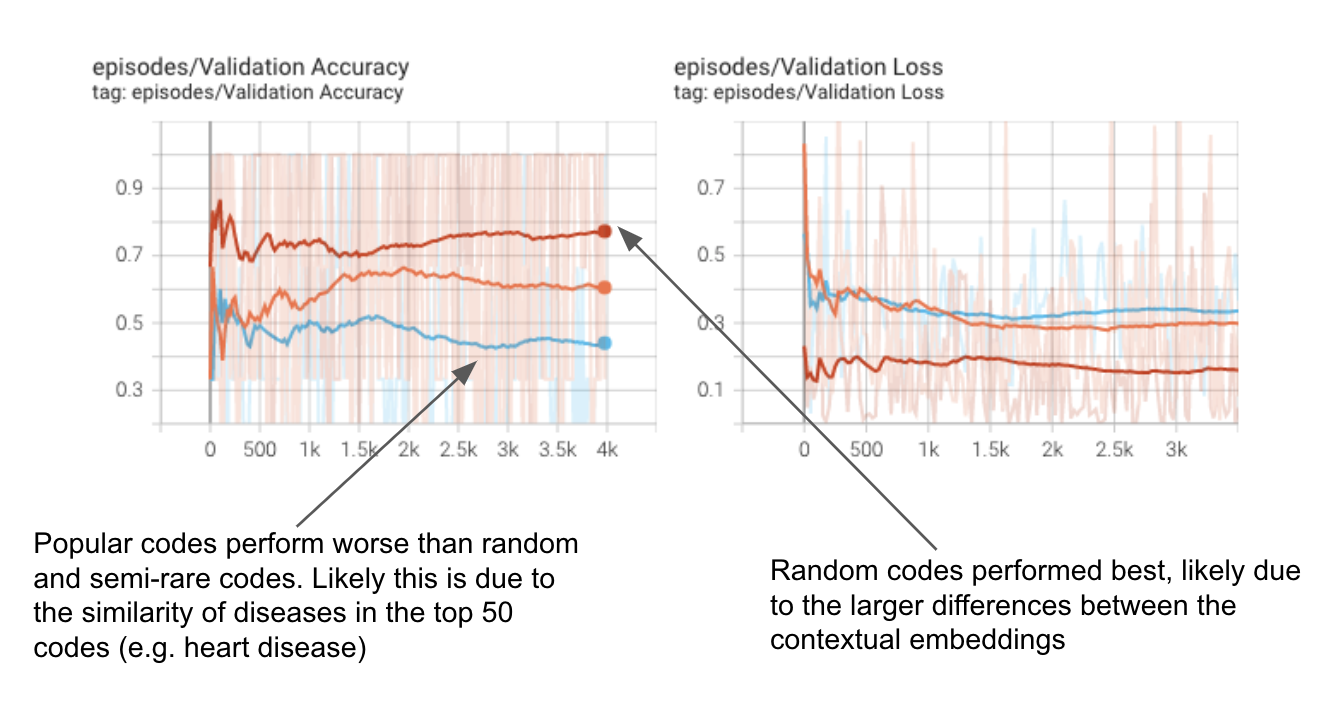}
  \caption{ProtoNet performance for random and semi-rare codes.}
  \label {fig: Protonet Rare}
\end{figure}

\subsection{Meta Learning with Distributionally Robust Optimization (DRO)}
Our purpose was to study the impact of predictions by meta learning models once they are optimized for worst case expected loss due to atypical groups of data. We used four variations of two meta-learning models each to compare test time performance. 
\begin{enumerate}
    \item Baseline meta-learning model: MAML and Protonet
    \item Model with DRO
    \item Model with Count Based or Group Adjusted DRO
    \item Model with Count Based or Group Adjusted DRO and $l_2$ regularization
\end{enumerate}
For our experiments disease codes were considered groupings with the assumption that there are distribution shifts due to temporal (written over 10 years) and spatial (written by different individuals within each group) shifts in the medical notes. We introduce the notation G which is a set of all the disease codes and g for the codes that are being used in the adaptation loop.
When sampling tasks we sample the triplet of (BERT Embeddings, labels, disease codes) as compared to (BERT Embeddings, labels) with a baseline model. The disease codes are only used in the outer adaptation loop and ignored in the inner loop loss calculation. Batch sizes were limited to 16 which results in high variance in the graphs. We also used batch size of 64 with less variance but achieved similar results.\newline 
For MAML we apply DRO in the outer loop adaptation only. We perform gradient descent based on the max group loss. We did not implement DRO for the inner loop, hence we do not use $g$ notation for $D^{tr}$. We wanted to focus on how well does the model "adapt" to distribution shifts.
We also track the losses and counts per group globally and use it to determine the worst case and best case disease codes and determine the performance during test time. The count is also used for group adjusted DRO.
We first established a baseline with CLINC150 (Figure: \ref{fig: DRO MAML Baseline}) and then used the MIMIC-III dataset for final results (Figure: \ref{fig: DRO MAML}).

The following is the DRO objective without group adjustment (not count based) used with MAML ( Finn et al., 2017 \cite{finn2017modelagnostic} and Sagaw et al., 2020 \cite{dronnforgroupshifts})
\begin{equation}
    \min_{ \theta } \max_{g \in G} \bigg\{ \sum_{task i} L(\theta - \nabla_{\theta}L((\theta, D^{tr}),D_{g}^{ts})) \bigg\}
\end{equation}

The following is the DRO objective with group adjustment (count based) used with MAML. The count term $n_g$ acts as a regularizer.( Finn et al., 2017 \cite{finn2017modelagnostic} and Sagaw et al., 2020 \cite{dronnforgroupshifts})
\begin{equation}
    \min_{ \theta }  \max_{g \in G} \bigg\{ \sum_{task i} L(\theta - \nabla_{\theta}L((\theta, D^{tr}),D_{g}^{ts})) + \frac{1}{\sqrt{n_g}} \bigg\}
\end{equation}

The following is the DRO objective without group adjustment (count based) used with ProtoNet. The count term $n_g$ acts as a regularizer.( Snell et al., 2017 \cite{snell2017prototypical}) and Sagaw et al., 2020 \cite{dronnforgroupshifts})
\begin{equation}
    \min_{ \theta }  \max_{g \in G} \bigg\{ \sum_{task i} L(\theta, D_{g}^{ts}) \bigg\}
\end{equation}

The following is the DRO objective with group adjustment (count based) used with ProtoNet. The count term $n_g$ acts as a regularizer.( Snell et al., 2017 \cite{snell2017prototypical}) and Sagaw et al., 2020 \cite{dronnforgroupshifts})
\begin{equation}
    \min_{ \theta }  \max_{g \in G} \bigg\{ \sum_{task i} L(\theta, D_{g}^{ts}) + \frac{1}{\sqrt{n_g}} \bigg\}
\end{equation}

\begin{figure}[H]
  \includegraphics[width=0.9\textwidth]{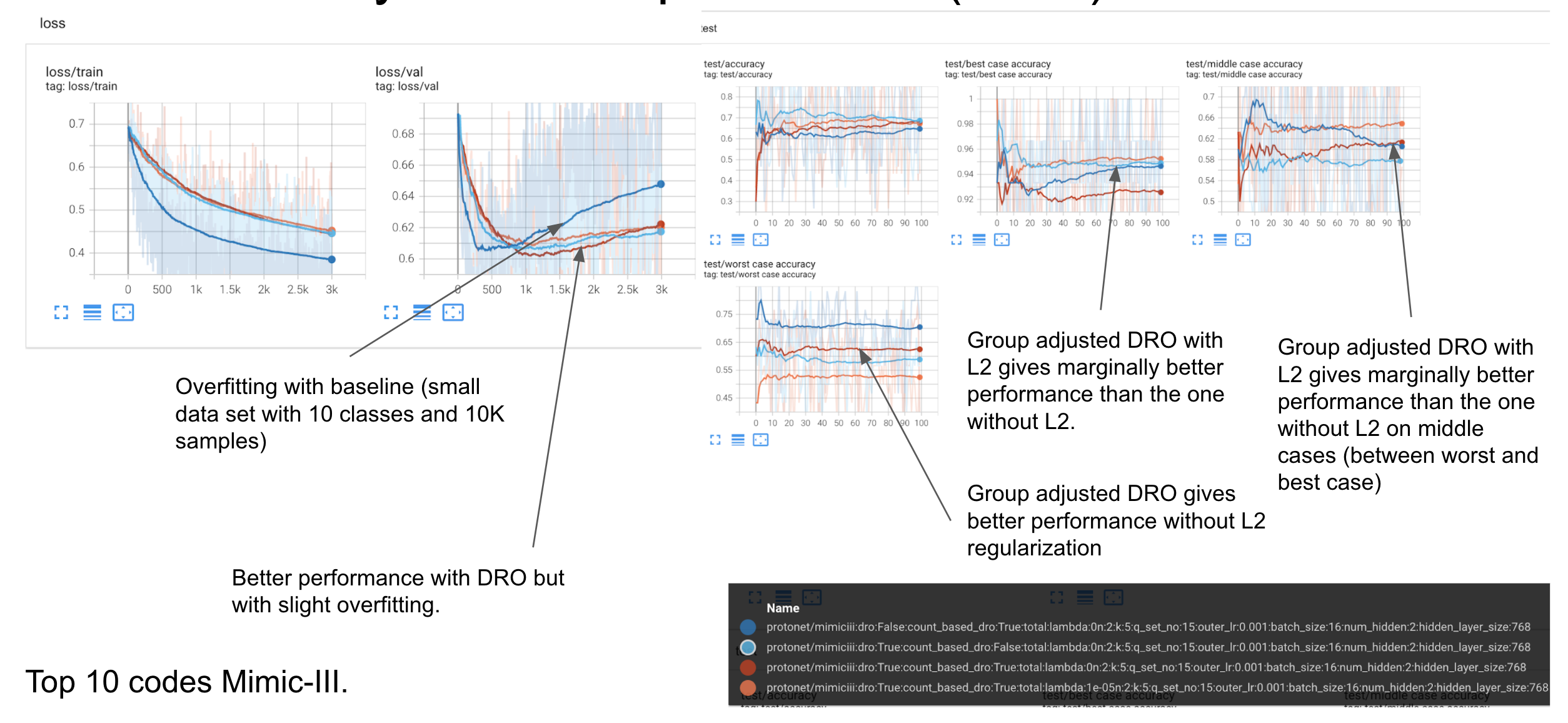}
  \caption{DRO with ProtoNet 2 class 5 shot model using top 10 MIMIC-III disease codes.}
  \label {fig: DRO Protonet}
\end{figure}

\textbf{Results} \newline
\begin{enumerate}
\item \textbf{MAML with DRO using CLINC150}\newline During test time, with CLINC150 we saw improved performance on worst case groups but slightly decreased accuracy for best case groups. This is expected as DRO also acts as a regularizer that forces the model to pay attention to worst case groups. See Figure \ref{fig: DRO MAML Baseline} for a summary of results. 

\begin{figure}[H]
  \includegraphics[width=0.9\textwidth]{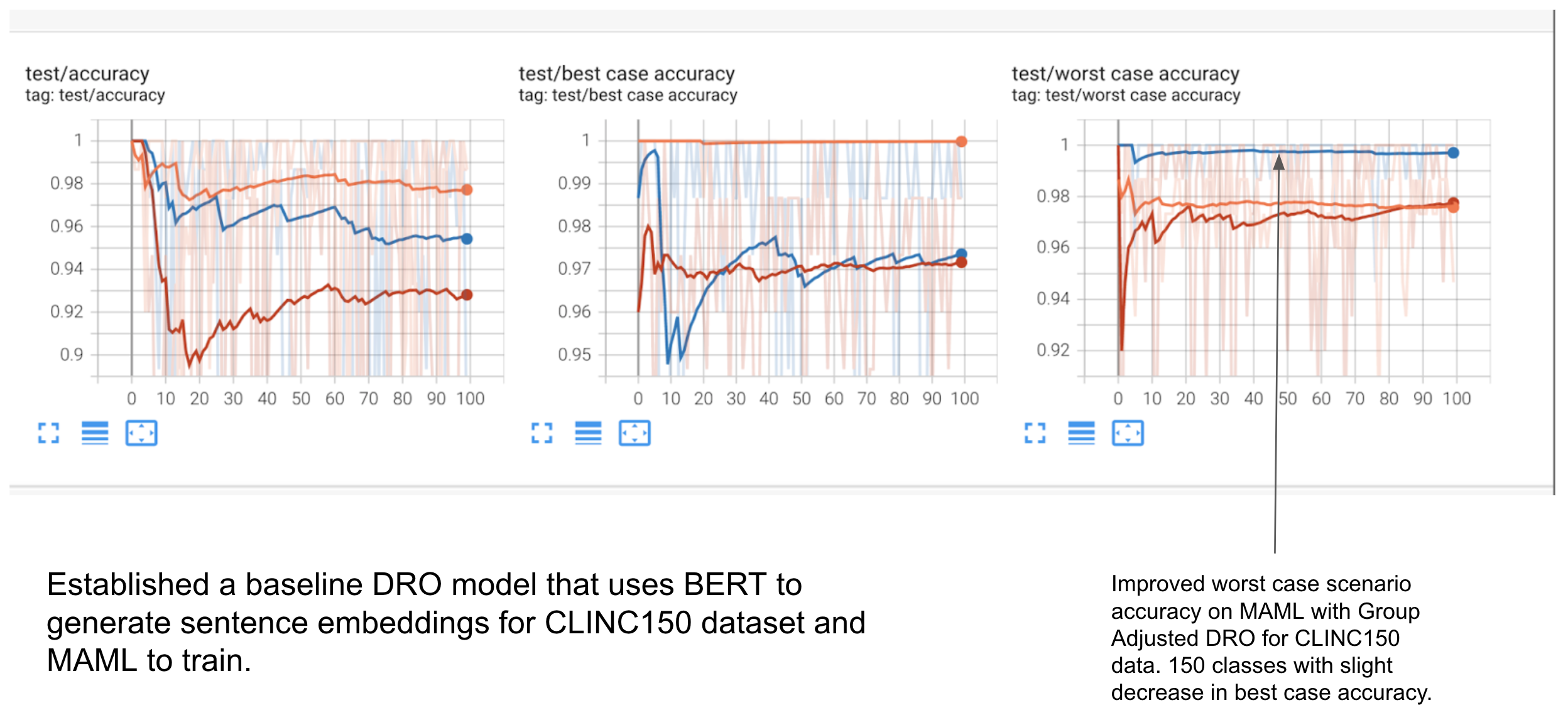}
  \caption{DRO Baseline with MAML using CLINC150 for a 150 class 3 shot model.}
  \label {fig: DRO MAML Baseline}
\end{figure}

\item \textbf{MAML with DRO using MIMICIII}\newline Please see Table \ref{table: MAML Results} and Figure \ref{fig: DRO MAML} for a summary of results. Baseline MAML showed signs of overfitting with the MIMICIII dataset. It did not perform well on the worst case groups. When DRO was enabled, there was no overfitting even when the $l_2$ regularizer was not used. With DRO there is drop in the best case accuracy but a perceptible increase or comparable results in the worst case scenarios. MAML with Count Based DRO (without $l_2$) performs well in comparison to other DROs as it show the most improvement on worst case and middle case groups. MAML with Count Based DRO (with $l_2$) performs better amongst all the DROs in best case scenarios.

\begin{table}[H]
\begin{tabular}{|c|c|c|c|c|}
\hline
Type & Avg & Worst Case & Best Case & Middle Case \\
\hline
Baseline MAML & 0.714$\pm$0.029 & 0.594$\pm$0.020 & \textbf{0.963$\pm$0.007} & 0.631$\pm$0.021 \\  
MAML DRO & \textbf{0.725$\pm$0.029} & 0.587$\pm$0.020 & 0.892$\pm$0.011 & 0.659$\pm$0.024 \\  
MAML Group Adjusted DRO (No $l_2$) & 0.696$\pm$0.031 & \textbf{0.646$\pm$0.018} & 0.908$\pm$0.010 & \textbf{0.698$\pm$0.030} \\  
MAML Group Adjusted DRO (With $l_2$) & 0.694$\pm$0.029 & 0.626$\pm$0.021 & 0.949$\pm$0.007 & 0.658$\pm$0.023 \\ 
\hline
\end{tabular}
\caption{DRO with MAML table of results. MAML with count based DRO and $l_2$ regularization seems to give less variance in results between Avg, Middle Case and Worst Case. Baseline MAML overfits the data and DRO acts as a regularizer.}
\label{table: MAML Results}
\end{table}

\begin{figure}[H]
  \includegraphics[width=0.9\textwidth]{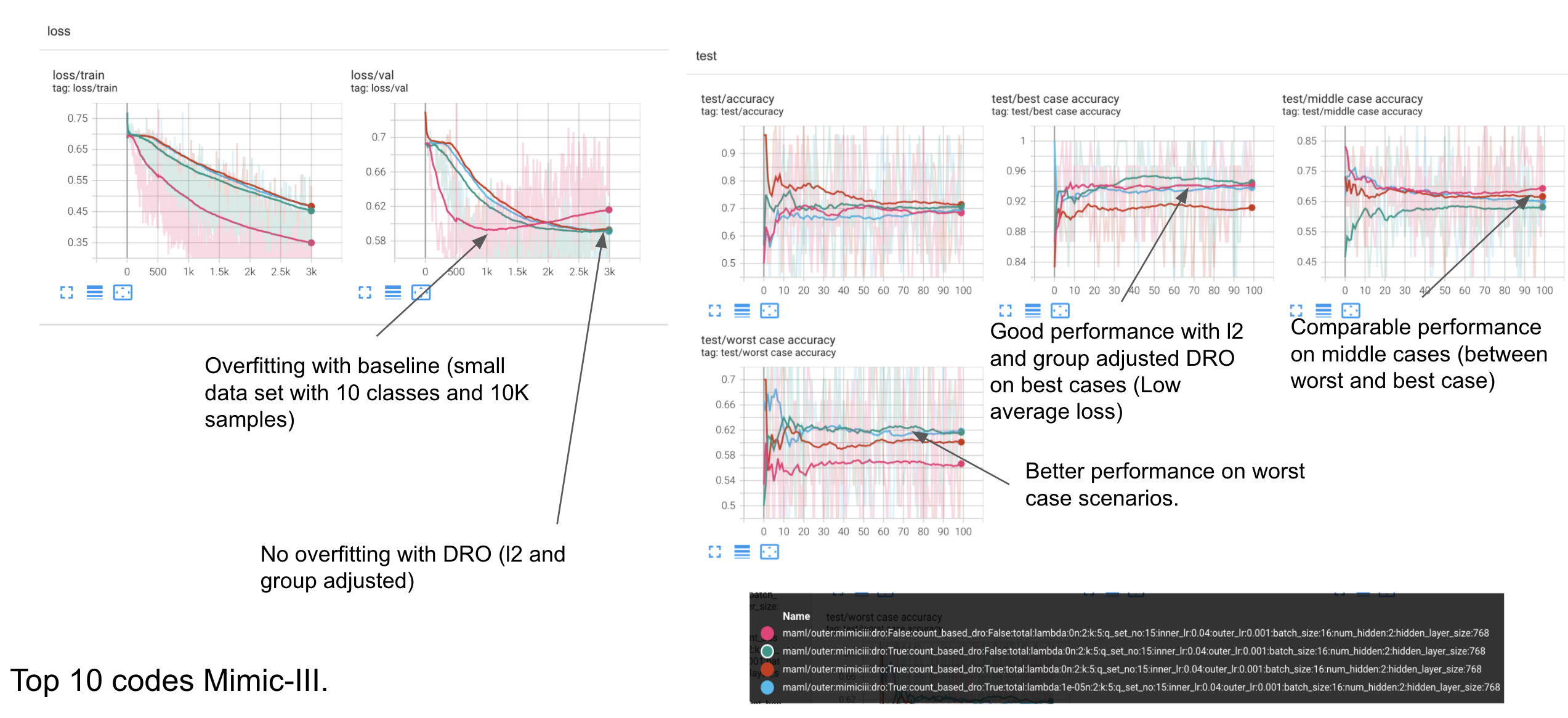}
  \caption{DRO in the outer loop with MAML 2 class 5 shot model using top 10 MIMIC-III disease codes.}
  \label {fig: DRO MAML}
\end{figure}

\item \textbf{ProtoNet with DRO using MIMICIII}\newline Please see Figure \ref{fig: DRO Protonet} for a summary of results. As with MAML, baseline ProtoNet overfits the MIMICIII data. DRO does regularize and reduce overfitting but does not eliminate it. Count Based or Group Adjusted DRO with $l_2$ regularization performs the best in all cases. For our experiment we passed the BERT embeddings through fully connected layers since we did not train BERT.
\end{enumerate}

\subsection{Prototypical Networks and Rare and Popular Disease Codes}
We compare the performance of ProtoNets on Clinical BERT embedding on random, semi-rare,
and popular disease classifications (Figure \ref{fig: Protonet Rare})
\begin{itemize}
    \item \textbf{Random:} Charts sampled from the full distribution of codes with greater than 10 notes per code
    \item \textbf{Semi-rare:} Charts sampled with codes with the bottom 50 codes with greater than 10 notes per code
    \item \textbf{Popular:} Charts among the top 50 codes (See Figure \ref{fig: All ICD Codes}).

\end{itemize}

\begin{table}[H]
\centering
\begin{tabular}{|c|c|c|}
\hline
n=10, k=5, q=1 & Average Meta-Test Accuracy & Meta Test STD \\ 
\hline
Popular & 45.70\% & 0.323 \\
Semi-rare & 62.40\% & 0.325 \\  
Random & 73.45\% & 0.238 \\
\hline
\end{tabular}
\caption{Meta Test Accuracy and standard deviation on popular, semi-rare and random ICD disease codes.}
\label{table: Rare and Popular Disease Codes}
\end{table}

\textbf{Results} \newline
Surprisingly, randomly distributed codes performed better than the popular and semi-rare codes. Likely the popular codes performed worse due to the skewed distribution of medical codes within the top codes which are focused on heart and lung diseases. This introduces a lower variation in the embeddings, making it difficult for the ProtoNet prototypes to be distinguished via minimum $l_2$ distance. Still, the meta-learning approach on rare codes has shown positive results. In contrast, fine-tuning a BERT model for similar performance required >500 samples, indicating that meta-learning on this dataset is significantly more memory efficient and useful for rare disease settings. 

\subsection{Discussion and Future Work}
Our purpose was to understand the performance of Meta-Learning in tackling low-resource  classification such as disease code classification using medical notes. We validated our approach on a NLP benchmark and devised two experiments on a medical dataset by coupling DRO with meta-learning algorithms and using ProtoNet to classify rare and popular disease codes. \newline

In our experiment, we did not exploit temporal data such as time-series data of disease change or all notes of a patient over each time period. This temporal data would be very useful to track the progress of each patient and get insight about the conditions development. In the future, processing each patient records as a whole observation would capture the time factor and would allow us to cover a broader spectrum of ICD codes. Another useful improvement could be the relationship between diseases as many diseases are correlated or even belong to a branch of a broader category. Exploiting the temporal data and hierarchy structure of ICD and integrating this information into our current baseline would improve model performance as beside comparing popular and rare diseases, we can also use top-down approach to compare each branch of disease.
\newline

We conclude that DRO combined with MAML does improve prediction and does accounts for distribution shift. The choice between DROs with or without $l_2$ may end up being domain specific. DRO combined with ProtoNet gives mixed results. We equalized the distribution of medical notes by choosing 1000 notes per disease code for sampling so that the distribution shift is limited to the content of the medical notes. Future experiments would involve removing that constraint (equalized distribution) and evaluate few shot learning on rare disease codes. \newline

In our architecture, BERT was used an embedding generator and was not trained using gradient descent. Future work could involve integrating the BERT loss function with meta-learning algorithms. Furthermore, we can investigate this approach on partial snippets of medical text to understand the limits of NLP meta-learning on these datasets. 
\newline

The results for ProtoNet predictions for random, popular and semi-rare codes are also encouraging. We believe that more experimentation and fine-tuning our encoder will give improved results.

\section{Related Works}

Zhang et al. (2019 \cite{zhang2019metapred}) used meta-learning on predicting risk using limited Electronic Health Records. They developed a model agnostic gradient framework to train a meta learner on a set of prediction tasks for relevant high risk clinical tasks. We wanted to adapt this concept using Prototypical Networks (ProtoNets)(Snell et al., 2017 \cite{snell2017prototypical}) and Model Agnostic Meta Learning - MAML (Finn et al., 2017 \cite{finn2017modelagnostic}). To account for distribution shifts in test data, Sagawa et al. (2020 \cite{dronnforgroupshifts}) achieved higher worst case accuracy by coupling DRO models with increased regularization. This forms the basis of our experiments with meta-learning where we combine DRO with MAML and ProtoNets to see if we can get similar results for Electronic Health Records using MIMICIII dataset. We also used Publicly available BERT embeddings (Alsentzer et al., 2019 \cite{alsentzer2019publicly}) to generate embeddings for the medical notes in the MIMICIII dataset. For data processing, we used the ideas of a data pipeline from MIMIC-Extract (Wang et al., 2020 \cite{2020mimicextract}) and NLP of MIMIC-III clinical notes (Nuthakki et al., 2019 \cite{nuthakki2019natural}) to build our own procedure of extracting and process related data before the embeddings generation. 

\section*{Team Contributions}

Our team's original breakdown of work for one individual to write the paper (Imran), one individual to manage the experiment data centrally (Pankaj), and one individual to research promising experiments (Minh). However, as we went through the research process, we found more effort was spend on designing and validating experiments, so we each contributed equal amounts to writing, designing, and managing data.

\section {Acknowledgements}
We are grateful to CS330 teaching staff for providing an engaging course with material that represents a frontier of machine learning!

\onecolumn
\bibliography{IEEEabrv,bibliography}

\onecolumn
\appendix

\end{document}